\documentclass[letterpaper,10pt,conference]{ieeeconf}
\IEEEoverridecommandlockouts       
\usepackage[ruled,vlined,linesnumbered]{algorithm2e}
\usepackage{float}
\usepackage{graphicx}
\usepackage{authblk}
\usepackage{amsmath}
\usepackage{amssymb}
\usepackage{wrapfig}
\usepackage{booktabs}
\usepackage{multirow}
\usepackage[table]{xcolor}
\usepackage[pdfa,colorlinks,bookmarksopen,bookmarksnumbered,allcolors=blue]{hyperref}
\SetKw{Return}{return}
\SetKw{KwContinue}{continue}
\SetAlFnt{\footnotesize}
\SetAlCapFnt{\footnotesize}
\SetAlCapNameFnt{\footnotesize}
\SetAlgoNlRelativeSize{-1}
\SetAlgoInsideSkip{smallskip}
\DontPrintSemicolon

\title{\LARGE \bf Foundation-Model-Guided Topology-Aware\\
Semantic Risk Fields for Manipulation}

\definecolor{ricemediumblue}{HTML}{4D9AD4}
\usepackage[
maxbibnames=99,
maxcitenames=2,
natbib=true,
style=ieee,
backend=biber,
sorting=none,
giveninits=true,
url=false, 
doi=false,
eprint=false,
isbn=false,
]{biblatex}

\author{Giung Lee, Weihang Guo, Lydia E. Kavraki%
\thanks{
GL, WG, LEK are with the Department of Computer Science, Rice University, Houston, TX, USA {\tt\small \{gl34, wg25, kavraki\}@rice.edu}. LEK is also affiliated with the Ken Kennedy Institute at Rice University. 
}
}

\begin{document}
\maketitle

\begin{abstract}
    Robot motion planning in everyday environments must satisfy hard geometric constraints while accounting for context-dependent semantic risk. We present a foundation-model-guided, topology-aware semantic risk field that extends manipulation safety beyond collision avoidance. For each manipulated-object/scene-object pair, a foundation model provides six directional risk weights and a pair-specific spatial decay scale. The method combines these priors with voxelized 3D scene geometry using topology-aware shielding and geodesic spatial decay. A GPU-parallel backend batches object-level distance and risk computations to construct a dense 3D field that serves as a modular cost for downstream motion planning. We evaluate the field's shielding behavior under full and partial barriers and compare its 3D workspace representation with a pixel-wise semantic-prior baseline. Across three household simulation scenarios, trajectories optimized with the proposed field have lower semantic exposure than collision-only trajectories under the same geometric constraints. We also evaluate foundation-model prior consistency and computational practicality. Together, these results support the proposed field as a practical topology-aware semantic cost representation for manipulation planning beyond collision avoidance.
\end{abstract}

\section{Introduction}

\begin{figure}[h]
    \centering
    \setlength{\abovecaptionskip}{0pt}
    \includegraphics[width=1.0\linewidth]{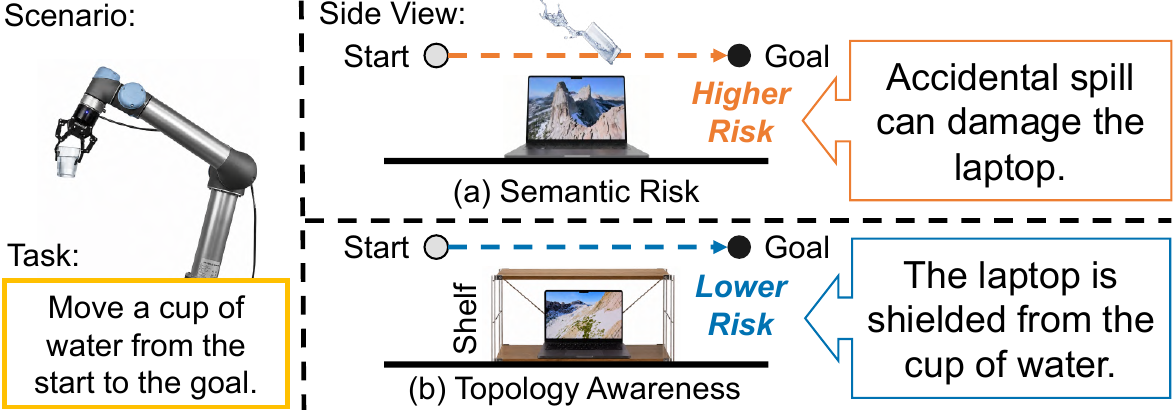}
    \caption{Motivating example of semantic and topology-aware manipulation safety with a UR5 carrying a cup of water. \textbf{(a)} Carrying water above a laptop is collision-free but semantically risky. \textbf{(b)} An intervening shelf shields the laptop, reducing the corresponding semantic risk.}
    \label{fig:sem_and_geo}
\end{figure}

Collision-free motion can still be semantically unsafe. Fig.~\ref{fig:sem_and_geo} illustrates this with a robot carrying a cup of water near a laptop. A path that moves directly above the laptop may satisfy all collision constraints and would therefore be considered safe by a collision-only planner, yet an accidental spill could damage the laptop.

More generally, robot safety is often implemented through geometric constraints, most prominently collision avoidance. Classical motion planners seek collision-free paths through configuration space using sampling-based methods such as PRM~\cite{kavraki1996probabilistic}, while trajectory-optimization methods such as CHOMP~\cite{zucker2013chomp} and TrajOpt~\cite{schulman2014motion} additionally optimize smoothness and related trajectory costs. However, collision avoidance is necessary but not sufficient for safe behavior in everyday environments. This paper focuses on a missing layer of safety: semantic risk in manipulation, where the safety cost depends on the manipulated object, nearby scene objects, and their spatial relationships.

Importantly, this risk is not determined solely by the semantic relationship between the manipulated object and the nearby scene object. It also depends on the surrounding environment's topology, including whether physical structures shield the scene object from potential consequences. In Fig.~\ref{fig:sem_and_geo} (b), an intervening shelf protects the laptop, making a trajectory above the shelf lower risk. This example motivates the need for a semantic risk representation that is both context-dependent and topology-aware, rather than based on collision geometry alone.

Recent work has expanded robot safety beyond geometric constraints by learning risk and safety specifications from demonstrations~\cite{li2022Learning, lindner2024learning} and by using foundation models to infer context-dependent hazards~\cite{brunke2025semantically, ravichandran2026contextual, santos2025updating}. Semantic-Metric Bayesian Risk Fields~\cite{chen2025semantic}, for example, infer risk from human videos with a VLM prior and output pixel-dense risk images conditioned on a query object. Although this representation captures spatially varying semantic risk in an image, it does not directly define a planner-ready risk field over the robot's 3D workspace. Moreover, it does not encode free-space connectivity or physical shielding: locations that are close in image or Euclidean space may nevertheless be separated by an obstacle.

Our contributions include: (i) a 3D directional semantic prior for pairs of manipulated and scene objects. A foundation model assigns risk scores along six fixed workspace directions ($\pm x$, $\pm y$, and $\pm z$) and estimates a pair-specific spatial scale $\sigma$ that controls spatial decay. We store these pairwise priors in a reusable database; (ii) a topology-aware 3D semantic risk voxel field that combines the priors with voxelized scene geometry obtained through RGB-D perception. The field propagates risk based on obstacle-aware geodesic distance rather than straight-line Euclidean distance and continuously attenuates risk when obstacles induce longer free-space detours; and (iii) a modular risk representation that can provide a cost for downstream robotic systems. In a case study of manipulation trajectory optimization in realistic household simulations, the proposed representation reduces semantic exposure relative to collision-only planning and captures topology-aware changes in risk propagation.

\section{Related Work}

\subsection{Semantic Safety with Foundation Models}
Traditional robot safety focuses primarily on physical constraints such as collision avoidance and joint and velocity limits. Semantic safety also considers object attributes, spatial relationships, common sense knowledge, and other contextual factors. Recent work~\cite{kim2026modular} has used foundation models to address semantic and contextual safety in robotic systems.

Several studies have examined semantic safety in manipulation. \citet{brunke2025semantically} combine language-model reasoning with a 3D semantic map to identify unsafe conditions. This method, however, relies on predefined spatial relations and does not account for how scene geometry affects risk propagation. Semantic-Metric Bayesian Risk Fields~\cite{chen2025semantic} generates query-conditioned, pixel-dense risk estimates from human videos and VLM priors, which can then be projected into 3D for downstream planning. It does not model topology-aware risk propagation or physical shielding. In mobile robotics and navigation, CORE~\cite{ravichandran2026contextual} and \citet{santos2025updating} ground foundation-model-derived safety constraints in 2D spatial representations, while RoboGuard~\cite{ravichandran2026guardrails} incorporates temporal-logic constraints for longer-horizon behavior.

\subsection{Topology-Aware Risk Representations}
Existing geometric and geodesic field methods represent spatial distance and connectivity without accounting for object-dependent semantic risk. Voxblox~\cite{oleynikova2017voxblox}, FIESTA~\cite{han2019fiesta}, and nvblox~\cite{millane2024nvblox} construct incremental 3D Euclidean signed distance fields from depth observations and are widely used for collision-aware planning. Their straight-line distances measure geometric proximity, but do not capture semantic relationships between objects or connectivity through collision-free space. A query point may therefore be close to a scene object in Euclidean distance even when reaching it requires a long collision-free detour. Riemannian and geodesic distance fields capture such non-Euclidean structure while preserving continuous field representations~\cite{li2024riemannian}, but they focus on geometric and dynamics-aware distance rather than semantic risk. \citet{cohn2025non} formulate non-Euclidean motion planning using graphs of geodesically convex sets, but similarly address geometric planning rather than object-dependent semantic risk. \citet{li2024csdf} construct geodesic flows over configuration-space distance fields that wrap around obstacles. These formulations distinguish locations that are close in Euclidean distance but separated by obstacles in the free-space topology.

Semantic grounding and geodesic-field methods are therefore complementary: the former determine which interactions are risky, whereas the latter determine how that risk should spread through the workspace.

\section{Problem Formulation}
\label{sec:problem_formulation}

Let $m$ denote the manipulated object, $\mathcal{O}$ the set of scene objects, and $\Omega\subset\mathbb{R}^3$ the bounded robot workspace. We use $E$ to represent the occupied environment geometry and $\Omega_{\mathrm{free}}(E)\subseteq\Omega$ to denote its collision-free subset. Our goal is to define a semantic risk cost over geometrically feasible workspace positions,
\begin{equation}
    V(m,p\mid E)
    =
    \sum_{o\in\mathcal O} V_o(m,p\mid E),
    \qquad
    p\in\Omega_{\mathrm{free}}(E),
    \label{eq:scene_risk_field}
\end{equation}
where $V_o(m,p\mid E)\geq0$ denotes the semantic risk induced by scene object $o$ when $m$ is placed at query position $p$ given the occupied geometry $E$. Occupied regions are handled separately by geometric collision constraints.

For downstream motion planning, let $\mathbf q$ denote a feasible robot joint trajectory and let $\mathbf p_r(\mathbf q)$ denote a task-dependent workspace reference point at which the field is queried. A risk-aware planner can incorporate the proposed field through an objective of the form
\begin{equation}
    \mathbf q^{*}
    =
    \arg\min_{\mathbf q}
    \left[
    J_{\mathrm{motion}}(\mathbf q)
    +
    \lambda
    \int_{\gamma_{\mathbf q}}
    V(m,p\mid E)\,ds
    \right],
    \label{eq:risk_aware_planning_objective}
\end{equation}
subject to the required start, goal, joint, and collision constraints. Here, $J_{\mathrm{motion}}(\mathbf q)$ denotes the conventional motion-planning objective independent of semantic risk, $\gamma_{\mathbf q}$ is the workspace path traced by $\mathbf p_r(\mathbf q)$, and $\lambda\geq0$ controls the contribution of semantic risk. Thus, geometric feasibility remains a hard requirement, while the proposed field $V$ provides an additional soft cost for reducing semantic exposure.

\section{Topology-Aware Semantic Risk Field}
\label{sec:semantic_risk_field}

\begin{figure*}[t]
    \centering
    \setlength{\abovecaptionskip}{0pt}
    \includegraphics[width=0.92\textwidth]{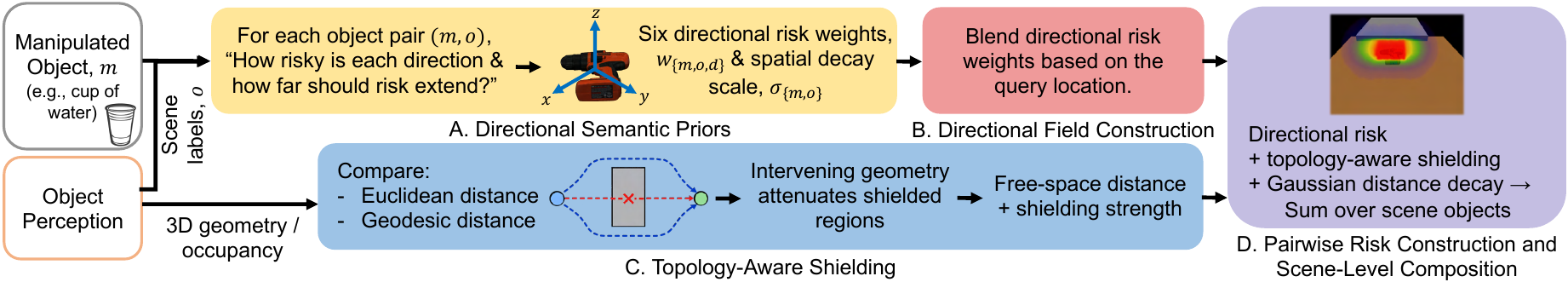}
    \caption{Overview of the proposed semantic risk field construction pipeline. Semantic priors for each manipulated-object/scene-object pair are combined with the observed 3D scene geometry to construct pairwise risk fields $V_o(m,p\mid E)$, which are aggregated into the scene-level field $V(m,p\mid E)$.}
    \label{fig:pipeline_overview}
\end{figure*}

Fig.~\ref{fig:pipeline_overview} summarizes the construction of the semantic risk field defined in Sec.~\ref{sec:problem_formulation}. The manipulated object $m$ is specified by the task, while each scene object is represented by a semantic label $o\in\mathcal O$ and its 3D geometry. These inputs may be provided by the object-perception frontend or supplied directly in controlled evaluations. For each ordered pair $(m,o)$, the system retrieves six directional weights $w_{m,o,d}$ and a pair-specific spatial decay scale $\sigma_{m,o}$. The observed 3D geometry and occupancy provide the directional and topology-dependent quantities that are combined with these priors to construct $V_o(m,p\mid E)$, and the pairwise fields are summed over $o\in\mathcal O$ to obtain $V(m,p\mid E)$. Because the object-level distance and risk-field computations are independent before summation into the final scene field, they can be batched and parallelized on a GPU.

For clarity, Table~\ref{tab:notation} summarizes the main notation used throughout this section.

\begin{table}[b]
\centering
\setlength{\abovecaptionskip}{0pt}
\caption{Notation used in the semantic risk field formulation.}
\label{tab:notation}
\scriptsize
\setlength{\tabcolsep}{2pt}
\renewcommand{\arraystretch}{0.95}
\begin{tabular}{@{}ll@{\hspace{5pt}}ll@{}}
\toprule
Symbol & Meaning & Symbol & Meaning \\
\midrule
$m$ & Manipulated object
& $o$ & Scene object \\

$\mathcal O$ & Scene-object set
& $p$ & Query position \\

$\Omega$ & 3D workspace
& $E$ & Occupied geometry \\

$\Omega_{\rm free}$ & Free workspace
& $B_o$ & Boundary seed voxels \\

$V$ & Scene risk
& $V_o$ & Object risk \\

$\mathcal D$ & Six directions
& $w_{m,o,d}$ & Directional weight \\

$\sigma_{m,o}$ & Spatial decay scale
& $\bar w_{m,o}$ & Blended directional risk \\

$d_{\rm euc}$ & Euclidean distance
& $d_{\rm geo}$ & Geodesic distance \\

$A_o$ & Topology attenuation
& $\phi$ & Risk mapping \\
\bottomrule
\end{tabular}
\end{table}

\subsection{Directional Semantic Priors}

A semantic risk prior for an object pair must capture both interaction type and direction. Prior semantic-safety work uses predefined spatial relations, such as \texttt{on}, \texttt{near}, and \texttt{around}~\cite{brunke2025semantically}, but these relations do not directly encode direction-dependent severity in the 3D workspace. For example, a single scalar per pair cannot distinguish carrying a heavy tool above a fragile glass from moving it beside the glass. Therefore, we represent each ordered pair $(m,o)$ by six directional weights in the workspace frame,
\begin{equation}
    w_{m,o,d}\in[0,1),
    \qquad
    d\in\mathcal D
    =\{\pm x,\pm y,\pm z\},\label{eq:directional_weight_domain}
\end{equation}
and a spatial decay scale $\sigma_{m,o}$.

\begin{wrapfigure}{r}{0.29\linewidth}
    \centering
    \setlength{\abovecaptionskip}{0pt}
    \vspace{-\baselineskip}
    \includegraphics[width=\linewidth]{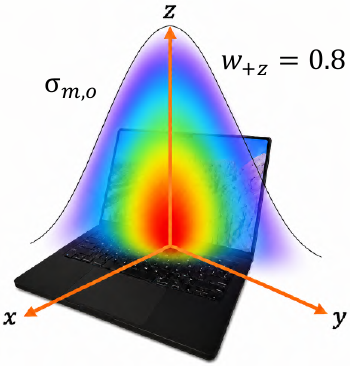}
    \caption{Each $w_{m,o,d}$ controls the direction-dependent semantic severity, while $\sigma_{m,o}$ controls the Gaussian decay of risk with geodesic distance.}
    \label{fig:dir_sem}
\end{wrapfigure}

The weight $w_{m,o,d}$ is a direction-conditioned semantic risk score that represents the severity of placing $m$ near $o$ along direction $d$; it is not interpreted as a probability. A value $w_{m,o,d}=0$ indicates that placing $m$ near $o$ along direction $d$ introduces no additional semantic risk. If all six weights are zero, we define $V_o(m, p\mid E)=0$ and do not use $\sigma_{m,o}$.

For a pair with nonzero semantic risk, $\sigma_{m,o}>0$ is a Gaussian spatial-decay scale measured in meters. Larger values spread risk over a broader free-space region, whereas smaller values localize it closer to object $o$.  We calibrate nonzero directional weights as low $(0,0.30)$, moderate $[0.30,0.70)$, high $[0.70,0.99)$, and operational near-no-go $[0.99,1)$, and use $0.03$--$0.20\,\mathrm{m}$ as the typical tabletop range for $\sigma_{m,o}$.

\subsection{Directional Field Construction}

We represent the workspace on a voxel grid to support efficient parallel field computation and direct integration with voxelized occupancy geometry. Because object $o$ itself occupies non-free voxels, we define $B_o$ as the set of free-space voxels immediately adjacent to its occupied boundary. We use $B_o$ as a common source set for both distance fields: $d_{\mathrm{euc}}(p,B_o)$ measures the shortest Euclidean distance from $p$ to $B_o$, while $d_{\mathrm{geo}}(p,B_o\mid E)$ measures the shortest collision-free workspace path length from $p$ to $B_o$. For each query position $p$, let $b_o(p)\in B_o$ denote the nearest source voxel and define the local offset $\Delta_o(p)=p-b_o(p)$. This offset describes the direction of $p$ relative to the object boundary. We use its signed coordinate components to determine how strongly each of the six workspace directions contributes at $p$. For coordinate index $j\in\{x,y,z\}$, the directional coefficients are
\begin{equation}
\begin{aligned}
    a_{o,+j}(p\mid E)
    &=
    \frac{\max(\Delta_{o,j}(p),0)}
    {\max(\|\Delta_o(p)\|_1,\varepsilon)},\\
    a_{o,-j}(p\mid E)
    &=
    \frac{\max(-\Delta_{o,j}(p),0)}
    {\max(\|\Delta_o(p)\|_1,\varepsilon)},
\end{aligned}
\label{eq:directional_coefficients}
\end{equation}
where $\varepsilon=10^{-6}$ prevents division by zero at a source voxel. We then linearly interpolate the raw directional weights as
\begin{equation}
    \overline w_{m,o}(p\mid E)
    =
    \sum_{d\in\mathcal D}
    a_{o,d}(p\mid E)w_{m,o,d}.
    \label{eq:interpolated_directional_weight}
\end{equation}
The linear interpolation preserves the $[0,1)$ prior scale and only affects how the six directional priors are blended at $p$. A query directly above $o$ inherits the $+z$ weight, while a diagonal query blends the corresponding signed-axis weights according to its offset.

\subsection{Topology-Aware Shielding}
\label{sec:Topology_Aware_Explaination}

Before accounting for the environment, the straight-line distance between $p$ and the object seed set is
\begin{equation}
    d_{\mathrm{euc}}(p,B_o)
    =
    \min_{b\in B_o}
    \|p-b\|_2.
    \label{eq:euclidean_distance}
\end{equation}
This distance does not distinguish a directly accessible point from one separated from the object by an obstacle. We therefore define the environment-conditioned geodesic distance
\begin{equation}
    d_{\mathrm{geo}}(p,B_o\mid E)
    =
    \min_{\gamma\in\Gamma(p,B_o\mid E)}
    \int_0^1\|\dot\gamma(t)\|_2\,dt,
    \label{eq:geodesic_distance}
\end{equation}
where $\Gamma(p,B_o\mid E)$ contains paths $\gamma:[0,1]\rightarrow\Omega_{\mathrm{free}}(E)$ satisfying $\gamma(0)=p$ and $\gamma(1)\in B_o$. If no free-space path exists, we set $d_{\mathrm{geo}}(p,B_o\mid E)=\infty$.

The topology-attenuation factor for object $o$ is
\begin{equation}
    A_o(p\mid E)
    =
    \operatorname{clip}\!\left(
    \frac{d_{\mathrm{euc}}(p,B_o)+\varepsilon}
         {d_{\mathrm{geo}}(p,B_o\mid E)+\varepsilon},
    0,1
    \right).
    \label{eq:geodesic_attenuation}
\end{equation}
When the object is directly accessible, $d_{\mathrm{euc}}\approx d_{\mathrm{geo}}$ and $A_o\approx1$. If an obstacle forces a detour, then $d_{\mathrm{geo}}>d_{\mathrm{euc}}$ and $0<A_o<1$. An unreachable query has $A_o=0$. The value of $A_o$ therefore varies continuously with the loss of direct accessibility caused by the topology of $E$ and does not set finite-detour regions exactly to zero.

\subsection{Pairwise Risk Construction and Scene-Level Composition}

Directly using bounded semantic weights cannot represent both graded ordinary risk and operational near-no-go risk, while infinite weights would make the objective discontinuous and unbounded. We therefore map the topology-attenuated effective severity $u=A_o(p\mid E)\,\overline w_{m,o}(p\mid E)\in[0,1)$ through the finite smooth function
\begin{equation}
    \phi(u)
    =
    \alpha u+(\Lambda-\alpha)u^\kappa,
    \qquad
    u\in[0,1)
    \label{eq:finite_risk_mapping}
\end{equation}
where $\alpha$ sets the ordinary-risk scale, $\kappa$ controls the near-one amplification, and $\Lambda$ sets its finite upper scale. Their experimental settings are given in Sec.~\ref{sec:experiments}. The same mapping $\phi$ is applied to every effective weight.

The upper calibration band represents operational near-no-go behavior rather than a mathematical exclusion constraint. Although the mapped field remains finite and decays with distance, the workspace $\Omega$ is bounded. Within the evaluated workspace, near-one weights therefore retain a sufficiently high cost to strongly discourage traversal without requiring an infinite weight.

The complete pairwise semantic risk field is: 
\begin{equation}
\boxed{
\begin{aligned}
V_o(m,p\mid E)
&=
\phi\!\left(
A_o(p\mid E)\,
\overline w_{m,o}(p\mid E)
\right)\\
&\quad\times
\exp\!\left(
-\frac{
d_{\mathrm{geo}}(p,B_o\mid E)^2
}{
2\sigma_{m,o}^2
}
\right).
\end{aligned}
}
\label{eq:pairwise_risk_field}
\end{equation}

We apply $A_o$ before \(\phi\) so that topology attenuates a near-one prior before nonlinear amplification. The geodesic distance then plays two complementary roles: $A_o$ captures relative detour, while the exponential term controls decay with absolute free-space distance. Sec.~\ref{sec:topology_validation} ablates these effects separately.

Finally, Eq.~\eqref{eq:scene_risk_field} sums the pairwise fields in Eq.~\eqref{eq:pairwise_risk_field}. The resulting $V(m,p\mid E)$ is queried along the workspace path $\gamma_{\mathbf q}$ in Eq.~\eqref{eq:risk_aware_planning_objective}, providing the semantic-cost term for planning.

\section{Experiments}
\label{sec:experiments}

Our experiments answer three questions:
\begin{itemize}
    \item Does the proposed method capture semantic risk and correctly reflect scene topology? Sec.~\ref{sec:semantic_representation} evaluates the proposed 3D representation against a pixel-wise semantic-prior baseline, and Sec.~\ref{sec:topology_validation} evaluates topology-aware shielding under full and partial barriers.

    \item Can the proposed field guide a downstream planner toward trajectories with lower semantic exposure under the same geometric constraints? Sec.~\ref{sec:risk_aware_planning} compares risk-aware and collision-only trajectories.

    \item Are the foundation-model priors reliable, and can the field and its supporting components be computed efficiently for practical manipulation workloads? Sec.~\ref{sec:prior_reliability} evaluates prior consistency across model families, while Sec.~\ref{sec:computational_performance} evaluates risk-field construction, perception, and prior-retrieval runtime.
\end{itemize}
All experiments are conducted on a Linux workstation with an Intel i7-12700K CPU and an NVIDIA RTX 4090 GPU with 24~GB of memory. Unless otherwise stated, we use $\alpha=50$, $\Lambda=10^6$, and $\kappa=200$ in Eq.~\eqref{eq:finite_risk_mapping} for all experiments. Here, $\alpha$ sets the ordinary-risk scale, $\kappa$ controls how sharply the nonlinear amplification is concentrated near the upper end of $[0,1)$, and $\Lambda$ specifies the finite limiting magnitude.

\subsection{Semantic Risk Representation}
\label{sec:semantic_representation}

We use OOPSIEVERSE~\cite{balaji2026oopsieverse} based on RoboCasa~\cite{robocasa2024} as household simulation testbeds. OOPSIEVERSE provides damage-aware manipulation environments built on realistic kitchen and household scenes. We replace the open-vocabulary perception stage with simulator-provided object poses and manually assigned semantic labels to isolate the spatial representation from perception and risk-prior generation.

\begin{figure}[h]
    \centering
    \includegraphics[width=1\linewidth]{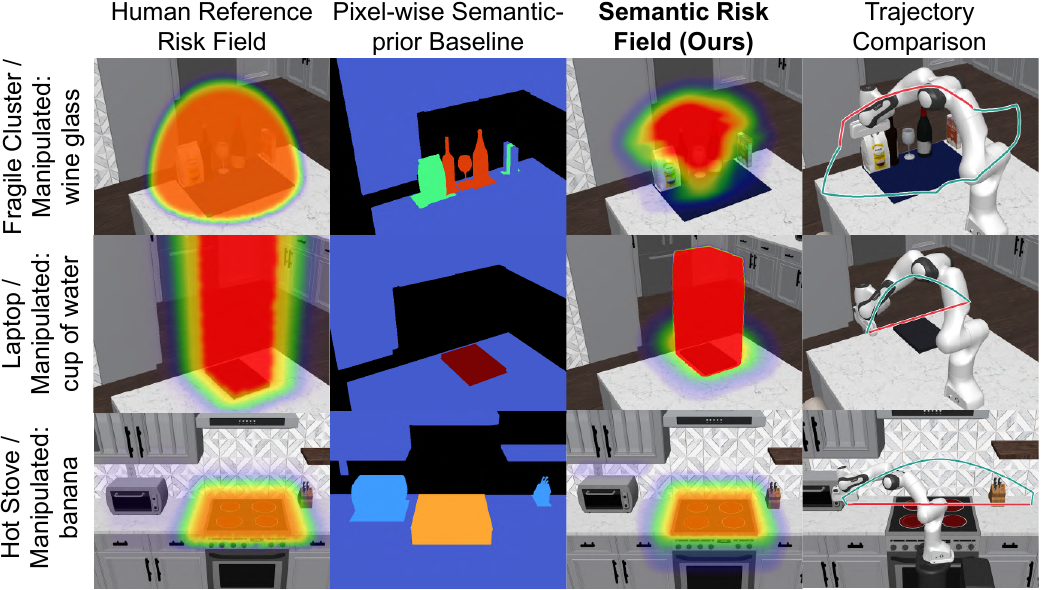}
    \caption{Oracle-labeled OOPSIEVERSE case studies for the fragile-cluster, laptop, and hot-stove scenarios. For each scenario, the columns show the human reference risk field, the pixel-wise semantic-prior baseline, the proposed 3D semantic risk field projected into the camera view, and the CHOMP trajectory comparison. The human reference risk fields are used only for evaluation and are not provided to the planners. Field colors range from higher risk in red/orange to lower risk in blue/violet. The red trajectory represents a collision-only path, and the green trajectory represents a risk-aware path.}
    \label{fig:robocasa_planning_cases}
\end{figure}

We consider three targeted OOPSIEVERSE kitchen scenarios, shown in Fig.~\ref{fig:robocasa_planning_cases}. In the first scenario, the robot carries a wine glass through a fragile-object cluster containing two wine bottles, a scene wine glass, a flour bag, and a cereal box. In the second scenario, the robot carries a cup of water near a laptop, where the primary hazard is not physical collision but potential spill exposure above the laptop. In the third scenario, the robot carries a banana near a hot stove. These scenarios respectively test risk arising from proximity to fragile objects, directional exposure to a spill-sensitive object, and proximity to a spatially extended heat source. For evaluation, we define a separate human reference risk field for each scenario: a vertically elongated half-ellipsoidal region around the fragile cluster, a rectangular region extending above the laptop, and a distance-decaying region around the hot stove. These fields are used only for evaluation and are hidden from both the proposed method and the planner.

We first compare our representation with the semantic-prior component of Semantic-Metric Bayesian Risk Fields~\cite{chen2025semantic}. The full method combines a semantics-driven prior with a learned, distance-conditioned likelihood. We isolate the prior to compare how the same semantic risk information is represented. For each scenario, we condition the baseline on the manipulated object. Simulator-provided segmentation assigns each visible reference-object pixel the same risk weight as in the human reference field in Fig.~\ref{fig:robocasa_planning_cases}. Walls, floors, and other structural background regions receive zero risk. The weights are in $[0,1)$, with values closer to one indicating greater risk. Both methods use the same oracle object identities and risk weights, so the comparison isolates the effect of spatial representation rather than perception or risk-prior generation.

Fig.~\ref{fig:robocasa_planning_cases} compares the resulting fields. Given the same object identities and risk weights, the pixel-wise semantic-prior baseline provides an image-space view of which visible object surfaces are risky. However, it does not describe how the risk extends into the surrounding 3D workspace through which the robot moves. Our semantic risk field instead assigns risk to world-aligned 3D voxels, identifying hazardous objects and nearby free-space regions that the robot should avoid. It can represent interaction-specific and directionally structured risk, such as the elevated spill-exposure region above the laptop and the spatial heat field around the stovetop, rather than limiting risk to visible object surfaces. The field can also be queried at arbitrary 3D positions.

\subsection{Topology-Aware Shielding Validation}
\label{sec:topology_validation}

We evaluate whether the proposed semantic risk field respects scene topology when physical geometry separates a risk source from a query region. We consider three full-shielding configurations: an overhead table above a laptop, a lateral wall beside a power drill, and a shelf containing a soccer ball. The first configuration evaluates vertical shielding, the second evaluates lateral shielding, and the third evaluates shielding within a more constrained multi-surface structure. We additionally test partial shielding by shifting the overhead table so that part of the laptop's upward risk remains uncovered and by shortening the side wall so that part of the power drill's lateral risk remains exposed.

For each scene, we fix the object pose, semantic prior, manipulated object, query position, and 1\,cm voxel resolution while removing only the shielding geometry. Thus, the straight-line source--query distance $d_{\mathrm{euc}}$ (Euclidean distance) remains fixed, whereas the shortest collision-free distance $d_{\mathrm{geo}}$ (geodesic distance) changes according to the available free-space paths. We report the topology-attenuation ratio $A=d_{\mathrm{euc}}/d_{\mathrm{geo}}$. Without an intervening obstacle, the two distances are equal and $A=1$. When a barrier forces a detour, $d_{\mathrm{geo}}$ becomes longer and $A<1$. Thus, a smaller value indicates stronger topology-aware shielding.

\begin{figure}[h]
    \centering
    \setlength{\abovecaptionskip}{0pt}
    \includegraphics[width=\columnwidth]{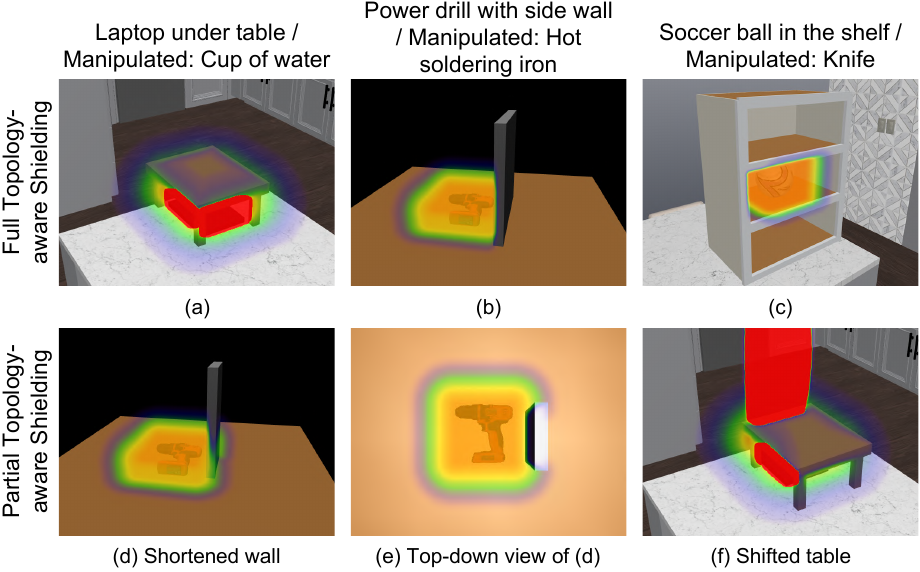}
    \caption{Topology-aware shielding under full and partial coverage. Full-shielding configurations are (a) an overhead table above a laptop with a cup of water manipulated, (b) a side wall beside a power drill with a hot soldering iron manipulated, and (c) a multi-rack shelf containing a soccer ball, with a knife as the manipulated object. Partial-shielding configurations shorten the side wall in (d) and shift the overhead table inward in (f). (e) is provided for better understanding of (d).}
    \label{fig:topology_shielding}
\end{figure}

Figure~\ref{fig:topology_shielding} shows how the proposed field changes when the intervening geometry increases the shortest free-space path between the source object and the query region. Under full shielding in Fig.~\ref{fig:topology_shielding} (a)--(c), the field is attenuated behind the overhead table, side wall, and shelf. The effect becomes strongest on the shelf, where the query is separated from the source by multiple surrounding surfaces. Under partial shielding in Fig.~\ref{fig:topology_shielding} (d) and (e), risk remains attenuated behind the shortened wall or shifted table but continues around the uncovered portion. This demonstrates that topology-aware shielding is spatially selective rather than an all-or-nothing visibility test.

Table~\ref{tab:topology_comparison} reports the full-shielding versus barrier-removed measurements at matched query positions. Removing the barrier gives $d_{\mathrm{geo}}=d_{\mathrm{euc}}$ and $A=1$ in all three scenes. With shielding geometry present, $d_{\mathrm{geo}}$ increases in every case, with the shelf producing the strongest relative detour.

\begin{table}[h]
\centering
\setlength{\abovecaptionskip}{0pt}
\caption{Counterfactual topology comparison at matched query positions. The straight-line distance is fixed at $d_{\mathrm{euc}}=0.230\,\mathrm{m}$. $A=d_{\mathrm{euc}}/d_{\mathrm{geo}}$ is the topology-attenuation ratio. Smaller values indicate stronger shielding.}
\label{tab:topology_comparison}
\scriptsize
\setlength{\tabcolsep}{9pt}
\begin{tabular}{lccc}
\toprule
Scene/condition
& $d_{\mathrm{euc}}$ (m)
& $d_{\mathrm{geo}}$ (m)
& $A$ \\
\midrule
\rowcolor{green!15}
Laptop: table present
& 0.230 & 0.395 & 0.582 \\
Laptop: table removed
& 0.230 & 0.230 & 1.000 \\
\addlinespace

\rowcolor{green!15}
Power drill: side wall present
& 0.230 & 0.351 & 0.655 \\
Power drill: side wall removed
& 0.230 & 0.230 & 1.000 \\
\addlinespace

\rowcolor{green!15}
Soccer ball: rack present
& 0.230 & 0.849 & 0.271 \\
Soccer ball: rack removed
& 0.230 & 0.230 & 1.000 \\
\bottomrule
\end{tabular}
\end{table}

The geodesic distance $d_{\mathrm{geo}}$ enters the field in two distinct ways: it determines the Gaussian distance-decay term and, together with $d_{\mathrm{euc}}$, the topology-attenuation ratio $A$. We isolate these two effects at the laptop query using three variants: (i) a Euclidean baseline with $d_{\mathrm{euc}}$ in the Gaussian decay term and topology attenuation disabled; (ii) geodesic-distance decay only, with $d_{\mathrm{geo}}$ in the Gaussian decay term and $A=1$; and (iii) the full formulation in Eq.~\eqref{eq:pairwise_risk_field}, with $d_{\mathrm{geo}}$ in the Gaussian decay term and $A=d_{\mathrm{euc}}/d_{\mathrm{geo}}$ for topology attenuation. All three variants use the same query, directional weight of $0.99999$, and spatial decay scale of $0.15\,\mathrm{m}$. We report the resulting risk-field value and its residual ratio. The residual ratio measures the fraction of the Euclidean-baseline risk remaining after each variant is applied, and is computed as the risk-field value of that variant divided by the Euclidean-baseline value. A value of $1$ indicates no reduction, whereas smaller values indicate stronger suppression.

\begin{table}[h]
\centering
\setlength{\abovecaptionskip}{0pt}
\caption{Component ablation at the matched laptop query.}
\label{tab:topology_component_ablation}
\scriptsize
\setlength{\tabcolsep}{5pt}
\begin{tabular}{lcc}
\toprule
Field construction
& Risk-field value
& Residual ratio \\
\midrule
Euclidean-distance decay only
& $3.080\times10^{5}$ & 1.000 \\
Geodesic-distance decay only
& $3.114\times10^{4}$ & 0.101 \\
\rowcolor{green!15}
Full topology-aware formulation
& $9.085\times10^{-1}$ & $2.949\times10^{-6}$ \\
\bottomrule
\end{tabular}
\end{table}

Table~\ref{tab:topology_component_ablation} shows that geodesic-distance decay alone yields a residual ratio of $0.101$, whereas the full formulation lowers the ratio to $2.949\times10^{-6}$. This difference reflects the distinct roles of the two components: geodesic distance captures absolute free-space separation, while $A$ further attenuates risk when intervening geometry makes the connection between the source and query indirect. Therefore, the intended topology-aware shielding behavior depends on both components.

\subsection{Risk-Aware Planning}
\label{sec:risk_aware_planning}

Using the same three household scenarios shown in Fig.~\ref{fig:robocasa_planning_cases}, we evaluate whether the proposed semantic risk field can guide manipulation trajectories toward lower semantic exposure under identical geometric constraints. For each scenario, we construct the proposed semantic risk field on a 1\,cm voxel grid and optimize a Panda joint-space trajectory using a deterministic multi-start CHOMP-style optimizer~\cite{zucker2013chomp}. We use PyRoki~\cite{kim2025pyroki} for forward kinematics and JAX~\cite{jax2018github} to differentiate the trajectory objective, with voxel-field risk sampled by trilinear interpolation. Both planner variants share the same start and goal states, initialization trajectories, occupancy information, and collision constraints. The collision-only objective favors shorter end-effector paths and includes orientation and joint-limit penalties along with a small joint-smoothness regularizer. The risk-aware objective also penalizes accumulated exposure to the proposed semantic field. We normalize each scene-level risk field before planning and use a fixed risk weight of \(\lambda=1000\) for all risk-aware experiments. In these experiments, we instantiate $p_r(q)$ in Eq.~\eqref{eq:risk_aware_planning_objective} as the Panda end-effector position and approximate the risk line integral using trajectory waypoints and joint-space midpoints. We use MuJoCo for collision checking.

We evaluate the resulting trajectories on the human reference risk fields defined in Sec.~\ref{sec:semantic_representation}. Their peak risks are $0.85$, $0.999$, and $0.80$ for the fragile-cluster, laptop, and hot-stove scenarios, respectively. We compute the mean and maximum path risks: mean risk is the average reference-field value over the sampled trajectory points, while maximum risk is the largest reference-field value encountered along the trajectory.

\begin{table}[h]
\centering
\setlength{\abovecaptionskip}{0pt}
\caption{CHOMP-planned EE trajectories evaluated on human reference risk fields.}
\label{tab:oracle_planning}
\scriptsize
\setlength{\tabcolsep}{3pt}
\begin{tabular}{llccc}
\toprule
Scenario & Planner & Mean path risk $\downarrow$
& Max risk $\downarrow$ & Path length (m) \\
\midrule
Fragile Cluster
& Collision-only
& 0.654 & 0.850 & 0.96 \\
\rowcolor{green!15}
Fragile Cluster
& \textbf{Risk-aware}
& \textbf{0.014} & \textbf{0.153} & 1.86 \\
\addlinespace

Laptop
& Collision-only
& 0.580 & 0.999 & 0.68 \\
\rowcolor{green!15}
Laptop
& \textbf{Risk-aware}
& \textbf{0.408} & \textbf{0.691} & 0.91 \\
\addlinespace

Hot Stove
& Collision-only
& 0.275 & 0.361 & 1.45 \\
\rowcolor{green!15}
Hot Stove
& \textbf{Risk-aware}
& \textbf{0.014} & \textbf{0.083} & 1.75 \\
\bottomrule
\end{tabular}
\end{table}

Table~\ref{tab:oracle_planning} shows why collision avoidance alone is insufficient. The laptop scene provides the clearest example: the collision-only trajectory passes directly above the laptop. Although this route is geometrically valid, it is dangerous when the robot carries a cup of water. The risk-aware planner accounts for the elevated spill-exposure region and instead detours around the laptop. The laptop scenario shows a smaller reduction in mean path risk because the human reference field extends broadly around the laptop, so the detouring trajectory still incurs some exposure. The fragile-cluster and hot-stove scenes similarly show risk-aware trajectories that avoid semantically hazardous regions, even when those regions do not constitute physical obstacles.

\subsection{Foundation-Model Prior Consistency}
\label{sec:prior_reliability}

We evaluated the consistency of the LLM fallback across model families using Claude Sonnet 4.6~\cite{anthropic2026sonnet46}, GPT-5.4-mini~\cite{openai2026gpt54mini}, and Qwen3-235B-A22B~\cite{qwen3technicalreport}. We queried each model once with the same finite-only prompt. All three models produced schema-valid JSON for each of the 135 ordered manipulated-object/scene-object entries, with six finite directional risk weights in $[0,1)$ and a valid spatial scale $\sigma_{m,o}$.

We then compared the generated priors against a human-rated prior reference for a stratified subset of 30 pairs. The subset included liquid, thermal, sharp, impact, contamination, structural-context, and ordered-pair cases, with 15 risky and 15 safe reference pairs. We classified a pair as risky when $\sigma_{m,o}>0$ and at least one directional weight was nonzero. Claude Sonnet 4.6, GPT-5.4-mini, and Qwen3-235B-A22B matched the reference for 27, 28, and 29 of the 30 pairs, respectively. We also evaluated whether each of the six directions was assigned zero or a nonzero risk. Across 180 directional decisions, the three models matched the reference in 162, 165, and 171 cases, respectively. Most remaining differences involved very low-risk assignments, along with one reversed ordered pair that all three models classified incorrectly. Qualitative agreement with the human-rated prior reference was similar across the three models.

\subsection{Computational Performance}
\label{sec:computational_performance}

\subsubsection{Risk-Field Construction Runtime}
\label{sec:risk_field_runtime}
We evaluate the backend runtime of semantic risk field construction as a function of scene clutter and voxel resolution. Our fully GPU-based backend batches objects and performs obstacle-aware geodesic distance computation, per-object risk-field evaluation, and scene-level superposition on-device in PyTorch, transferring only the final field to the host. Geodesic distances are computed using an iterative Godunov solver for the Eikonal equation~\cite{jeong2008fast}. Experiments are conducted in MuJoCo~\cite{todorov2012mujoco} scenes containing \(N \in \{1,2,4,6,8,10,20,50,100\}\) scene objects drawn from the YCB object dataset~\cite{calli2015ycb} and placed on the table. For each object count, we evaluate voxel resolutions of 8\,mm, 1\,cm, 2\,cm, and 4\,cm. The workspace measures $1\,\mathrm{m} \times 1\,\mathrm{m} \times 1\,\mathrm{m}$. Object poses are randomly sampled on the tabletop subject to non-overlap constraints, and the sampled layout for each object-count setting is held fixed across repeated runtime trials. The camera pose, workspace bounds, manipulated object, object labels, grounded object geometry, and semantic prior cache are also held fixed across resolutions, so that only the voxel resolution and number of scene objects vary.

Each object-count/resolution setting is repeated 100 times after one warm-up run, and we report mean runtime and standard deviation. Runtime is measured from perception output to the final dense scene-level field $V$, including voxel-grid construction, occupancy generation, Euclidean and geodesic distance computation, per-object risk-field construction, and scene-level superposition. Perception, visualization, and file-saving overhead are excluded.\vspace{-5pt}

\begin{table}[h]
\centering
\setlength{\abovecaptionskip}{0pt}
\caption{Full GPU backend runtime (in milliseconds). Each entry reports the mean $\pm$ standard deviation over 100 recorded runs after one unrecorded warm-up.}
\label{tab:runtime_grid}
\scriptsize
\setlength{\tabcolsep}{3pt}
\begin{tabular}{lcccc}
\toprule
\# Objects & 8\,mm & 1\,cm & 2\,cm & 4\,cm \\
\midrule
1   & $63.4 \pm 0.9$ & $40.0 \pm 0.9$ & $16.0 \pm 0.6$ & $12.2 \pm 0.7$ \\
2   & $82.3 \pm 1.4$ & $49.2 \pm 0.8$ & $17.7 \pm 0.7$ & $13.2 \pm 0.6$ \\
4   & $137.7 \pm 2.1$ & $80.3 \pm 1.6$ & $23.7 \pm 0.9$ & $15.9 \pm 0.6$ \\
6   & $177.9 \pm 2.8$ & $102.4 \pm 2.2$ & $28.6 \pm 0.9$ & $18.0 \pm 0.5$ \\
8   & $256.0 \pm 4.7$ & $142.3 \pm 3.8$ & $35.5 \pm 1.6$ & $20.1 \pm 0.7$ \\
10  & $265.0 \pm 4.9$ & $149.5 \pm 3.5$ & $36.8 \pm 0.9$ & $20.8 \pm 0.6$ \\
20  & $451.5 \pm 9.6$ & $247.0 \pm 6.9$ & $56.9 \pm 1.0$ & $29.3 \pm 0.8$ \\
50  & $960.4 \pm 24.1$ & $523.8 \pm 9.8$ & $115.8 \pm 2.2$ & $56.4 \pm 1.5$ \\
100 & $1759.9 \pm 50.7$ & $1017.7 \pm 16.0$ & $223.6 \pm 4.0$ & $116.9 \pm 3.5$ \\
\bottomrule
\end{tabular}
\end{table}

\begin{figure}[h]
    \centering
    \setlength{\abovecaptionskip}{0pt}
    \includegraphics[width=0.85\linewidth]{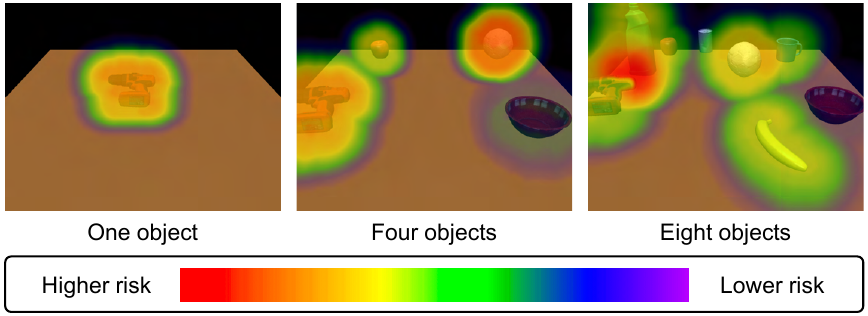}
    \caption{Semantic risk-field overlays for scenes in which a hot soldering iron is the manipulated object.}
    \label{fig:risk_fields_example}
\end{figure}

As shown in Table~\ref{tab:runtime_grid}, construction time scales approximately linearly with the number of scene objects and increases substantially at finer voxel resolutions. Such throughput is practical for typical manipulation workloads: at 2~cm resolution, the backend constructs a 20-object scene in 56.9~ms, corresponding to an update rate of approximately 18~Hz. The qualitative examples in Fig.~\ref{fig:risk_fields_example} show representative risk fields for sparse and cluttered tabletop scenes. As more semantically relevant objects are introduced, their individual risk fields are combined into the final scene-level field $V$. Regions in which multiple object-level fields overlap receive a higher risk under the superposition rule, reflecting locations where the manipulated object is simultaneously risky with respect to multiple scene objects.

\subsubsection{Perception Frontend Evaluation}
\label{sec:perception_frontend}

Our object-perception consists of two modular stages. Given an RGB observation, an automatic mask generator first produces a set of instance masks. SigLIP~2~\cite{tschannen2025siglip2} is then applied to the resulting masked regions to assign semantic object labels. We separately measure the mask-generation latency for the tabletop scene, containing $N \in \{1,2,4,6,8,10\}$ objects. We compare the automatic mask-generation modes of SAM~2.1 (Hiera-S)~\cite{ravi2024sam2}, MobileSAMv2~\cite{zhang2023mobilesamv2}, and FastSAM-s~\cite{zhao2023fastsam}. For each object count, we use one fixed $640 \times 480$ RGB observation and perform 100 recorded runs after one unrecorded warm-up run.

\begin{table}[h]
    \centering
    \setlength{\abovecaptionskip}{0pt}
    \caption{Runtime comparison of mask-generation methods. Latency is reported in milliseconds as mean $\pm$ standard deviation over 100 runs. $N$ is the number of objects.}
    \label{tab:mask_generator_runtime}
    \begin{tabular}{rccc}
        \toprule
        \textbf{$N$}
        & \textbf{SAM 2.1}
        & \textbf{MobileSAMv2}
        & \textbf{FastSAM-s} \\
        \midrule
        1  & $1276.5 \pm 25.6$ & $62.2 \pm 1.8$ & $7.6 \pm 1.1$ \\
        2  & $1296.8 \pm 26.0$ & $62.3 \pm 1.7$ & $7.0 \pm 0.4$ \\
        4  & $1346.4 \pm 43.8$ & $66.6 \pm 1.3$ & $7.5 \pm 0.3$ \\
        6  & $1273.2 \pm 21.5$ & $67.3 \pm 2.2$ & $8.5 \pm 0.6$ \\
        8  & $1267.2 \pm 28.5$ & $69.2 \pm 2.2$ & $8.7 \pm 0.7$ \\
        10 & $1238.6 \pm 77.9$ & $71.8 \pm 1.9$ & $8.6 \pm 0.6$ \\
        \bottomrule
    \end{tabular}
\end{table}

This comparison characterizes the trade-off between latency and mask quality in the modular perception frontend. Across six scenes containing 31 objects, SAM~2.1, MobileSAMv2, and FastSAM-s generated valid masks for 29, 30, and 31 objects, respectively, while spurious masks appeared in 0, 1, and 6 of the six scenes. In this evaluation, faster mask generation corresponded with spurious detections in more scenes, so the appropriate frontend depends on the application's requirements for latency and mask quality. Perception quality itself is outside the scope of this work. \vspace{-6pt}

\begin{figure}[h]
    \centering
    \setlength{\abovecaptionskip}{0pt}
    \includegraphics[width=0.9\linewidth]{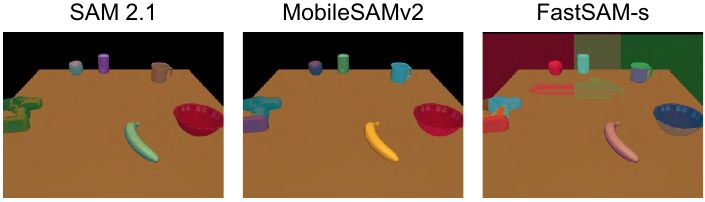}
    \caption{Mask-generation results using three different models on a tabletop scene containing 6 objects.}
    \label{fig:mask_generation_figure}
\end{figure} \vspace{-4pt}

\subsubsection{Prior Retrieval and Fallback Runtime}
\label{sec:prior_fallback_runtime}

We measured fallback overhead in the eight-object environment using Claude Sonnet~4.6 through the Anthropic API. To invoke the fallback, we withheld the prior for (\emph{hot soldering iron}, \emph{power drill}) and retained the other seven. We compared three conditions: all eight priors precomputed, one cold LLM fallback, and repeated retrieval after in-memory caching. Across 10 trials, retrieving all eight precomputed priors took less than $0.02$\,ms. The cold fallback took a median of $2.11$\,s, with an interquartile range of $2.01$--$2.20$\,s. Once the generated prior was cached, retrieving all eight priors took less than $0.01$\,ms. Therefore, the initial remote LLM query dominates the observed latency, while cached retrieval adds negligible overhead to subsequent field construction.

Secs.~\ref{sec:semantic_representation}--\ref{sec:computational_performance} address the three questions posed at the beginning of this section. The representation and shielding results show that the proposed field captures semantic structure and responds to scene topology, while the planning results show that it reduces semantic exposure under the same geometric constraints. The prior-consistency results show broad agreement across model families and the human-rated prior reference. The runtime results indicate practical computation: at $2$\,cm resolution, the GPU backend constructs a 20-object field in $56.9$\,ms, or approximately $18$\,Hz.

\section{Conclusion and Limitations}

We presented a foundation-model-guided, topology-aware semantic risk field that combines directional semantic priors with voxelized 3D geometry to reduce semantic exposure while capturing topology-aware shielding. The current evaluation is limited to static, near-field scenes and depends on perception, calibration, fixed-frame directional priors, and foundation-model prior quality. Future work will explore object-relative priors, dynamic environments, and mobile manipulation and navigation. We hope these results motivate robot-safety methods that account for semantic and context-dependent risk alongside collision avoidance.

\section*{Acknowledgment}
LEK and WG have been supported in part by NSF 2336612 and Rice University Funds

\printbibliography{}

\end{document}